\documentclass{article}
\usepackage{spconf,amsmath,graphicx,bbm, comment}
\usepackage{multirow, hhline, stmaryrd}
\usepackage{algorithm, algorithmic}
\DeclareMathOperator*{\argmax}{arg\,max}

\title{Conditional Teacher-Student Learning}
\name{Zhong Meng, Jinyu Li, Yong Zhao, Yifan Gong}
\address{Microsoft Corporation, Redmond, WA, USA} 

\begin{document}
\ninept
\maketitle
\begin{abstract}


The teacher-student (T/S) learning has been shown to be effective for a variety of problems such as domain adaptation and model compression.  One shortcoming of the T/S learning is that a teacher model, not always perfect, sporadically produces wrong guidance in form of posterior probabilities that misleads the student model towards a suboptimal performance. To overcome this problem, we propose a conditional T/S learning scheme, in which a ``smart'' student model selectively chooses to learn from either the teacher model or the ground truth labels \emph{conditioned on} whether the teacher can correctly predict the ground truth. Unlike a naive linear combination of the two knowledge sources, the conditional learning is exclusively engaged with the teacher model when the teacher model's prediction is correct, and otherwise backs off to the ground truth. Thus, the student model is able to learn effectively from the teacher and even potentially surpass the teacher. We examine the proposed learning scheme on two tasks: domain adaptation on CHiME-3 dataset and speaker adaptation on Microsoft short message dictation dataset. The proposed method achieves 9.8\% and 12.8\% relative word error rate reductions, respectively, over T/S learning for environment adaptation and speaker-independent model for speaker adaptation. 

\end{abstract}
\section{Introduction}
\label{sec:intro}
Teacher-student (T/S) learning \cite{li2014learning, hinton2015distilling}  has been widely applied to a variety of deep learning tasks in speech, language and image processing including model compression \cite{li2014learning, hinton2015distilling}, domain adaptation \cite{li2017large, meng2018adversarial, movsner2019improving}, small-footprint natural machine translation (NMT) \cite{kim2016sequence}, low-resource NMT \cite{chen2017teacher}, far-field automatic speech recognition (ASR) \cite{li2018developing, watanabe2017student}, low-resource language ASR \cite{cui2017knowledge} and neural network pre-training \cite{tang2016recurrent}.
T/S learning falls in the category of transfer learning, where the network of interest, as a student, is trained by mimicking the behavior of a well-trained network, as a  teacher, in the presence of the same or stereo training samples. Formally, the T/S learning works by minimizing the Kullback-Leibler (KL) divergence between the output distribution of the student and teacher models, other than from the hard labels derived from the transcriptions. 

Compared to using conventional one-hot hard label as the training
target, the transfer of soft posteriors \cite{li2014learning} well preserves the probabilistic relationships among different classes encoded at the output of the teacher model. Because soft labels provide more information than hard labels for the model training, the T/S learning results in better performance as reported in \cite{li2014learning, hinton2015distilling, li2018developing}. The largest benefits of using pure soft labels is learning without any hard labels, enabling the use of much larger amount of 
unlabeled data to improve the student model performance \cite{li2014learning,  li2018developing}.

One shortcoming of the T/S learning is that a teacher model, not always perfect, sporadically makes the incorrect predictions that mislead the student model towards a suboptimal performance. In such a case, it may be beneficial to utilize hard labels of the training data to alleviate this effect. Hinton et. al. \cite{hinton2015distilling}
later proposed an interpolated T/S learning called knowledge distillation, in which a weighted sum of the soft posteriors and the one-hot hard label is used to train the student model. 
One issue is that the simple linear combination with
one-hot vectors destroys the relationships among different
classes embedded naturally in the soft posteriors produced by the teacher
model. Moreover, proper setting of the interpolation weight with a fixed value is known to be critical and it varies with the adaptation scenarios and the qualities of the teacher and ground truth labels. 

In this paper, we propose a
\emph{conditional T/S learning} scheme, where the student model becomes smart so that it can criticize the  knowledge imparted by the teacher model to make better use of the teacher and the ground truth. 
At the initial stage, when the
student model is very weak, it blindly follows whatever knowledge infused by the
teacher model and uses the soft posteriors as the solely training targets. As
the student model grows stronger, it begins to selectively choose the
learning source from either the teacher model or the ground
truth labels \emph{conditioned on} whether the teacher's prediction coincides with the ground truth. That is, the student model would learn exclusively from the teacher when the teacher makes correct prediction
on training samples, and otherwise from the ground truth when the teacher is wrong. 
With conditional T/S learning, the student makes good use of rich and correct knowledge encompassed by the teacher, while avoids receiving inaccurate knowledge generated by the teacher. Another advantage of the conditional T/S learning over the conventional T/S learning is that it forgoes tuning the interpolation weight between two knowledge sources.

We applied the proposed approach to two tasks, domain adaptation and speaker adaptation. In domain adaptation, 
the student model is trained using the noise corrupted data in the target domain as input and the soft target obtained from the teacher posterior computed on the corresponding clean data. 
We demonstrate the effectiveness of the proposed approach using CHiME-3 dataset. 
In speaker adaptation, it can be shown that the conventional T/S learning is equivalent to the KLD adaptation \cite{asami_2017}, where the speaker-independent model acts as a teacher and the speaker dependent model acts as a student. Similarly, we apply the conditional T/S learning to further boost the performance of the KL divergence (KLD) adaptation. We demonstrate the improvement over the KLD adaptation for supervised and unsupervised adaptation on the Microsoft Windows Phone short message dictation task.
\section{Teacher-Student Learning}

\label{sec:ts}
In T/S learning, a well-trained teacher network  takes in an
sequence of training samples $\mathbf{X}^T=\{\mathbf{x}^T_{1}, \ldots,
\mathbf{x}^T_{N}\}, \mathbf{x}^T_i \in \mathbbm{R}^{D_T}$ and predicts a sequence of
class labels. Here, each class is represented by an integer $c \in \{1, 2,\ldots, D_C\}$ and $D_C$ is the total
number of classes in the classification task.  The goal is to learn a
student network that can accurately predict the class labels  for each of the its input samples
$\mathbf{X}^S=\{\mathbf{x}^S_{1}, \ldots, \mathbf{x}^S_{N}\}, \mathbf{x}^S_i \in
\mathbbm{R}^{D_S}$ by using the knowledge transferred from the teacher
network. To ensure effective knowledge transfer, the input sample sequences
$\mathbf{X}^T$ and $\mathbf{X}^S$ need to be parallel to each other, i.e, each pair of train
samples $\mathbf{x}^T_i$ and $\mathbf{x}^S_i$ share the same ground truth
class label $c_i\in \{1, 2, \ldots, D_C\}$.


\subsection{T/S Learning with Soft Labels}

T/S learning minimizes the Kullback-Leibler (KL) divergence
between the output distributions of the teacher network and the student network
given parallel data $X^T$ and $X^S$ are at the input to the networks \cite{li2014learning}.
The KL divergence between the teacher and student output distributions
$p(c|\mathbf{x}^T_i; \mathbf{\theta}_T)$ and $p(c|\mathbf{x}^S_i; \mathbf{\theta}_S)$
is formulated as:
\begin{align}
	&\mathcal{KL}\left[p(c|\mathbf{x}^T_i; \mathbf{\theta}_T)||p(c|\mathbf{x}^S_i; \mathbf{\theta}_S)\right] = \nonumber \\
	&\qquad\qquad\qquad \sum_{i=1}^N\sum_{c = 1}^{D_C}
	p(c|\mathbf{x}^T_i; \mathbf{\theta}_T) \log
	\left[ \frac{p(c|\mathbf{x}^T_i; \mathbf{\theta}_T)}{p(c |\mathbf{x}^S_i; \mathbf{\theta}_S)}
	\right], 
	\label{eqn:kld_ts}
\end{align}
 $i$ is the sample index, $\mathbf{\theta}_T$ and $\mathbf{\theta}_S$ are the parameters of the teacher
and student networks, respectively, $p(c|\mathbf{x}^T_i; \mathbf{\theta}_T)$ and $p(c|\mathbf{x}^S_i; \mathbf{\theta}_S)$ are the posteriors of class $c$ predicted by the teacher and student network given the input samples $\mathbf{x}^T_i$ and $\mathbf{x}^S_i$, respectively. To learn a student network that
approximates the given teacher network, we minimize the KL divergence with
respect to only the parameters of the student network while keeping the
parameters of the teacher network fixed, equivalent to minimizing
the loss function below:\footnote{In some cases, the senone posteriors generated by the teacher network are flattened by a temperature $T > 1$ before serving as the soft labels \cite{hinton2015distilling}. But in speech area, $T$ is normally fixed at $1$ \cite{watanabe2017student,chan2015transferring,tan2018knowledge}. We obtain the best performance when $T=1$ and the same conclusion is also reported in \cite{lu2017knowledge,kim2016sequence}.}
\begin{align}
	\mathcal{L}_{TS}(\mathbf{\theta}_S) =
	-\frac{1}{N}\sum_{i=1}^N\sum_{c = 1}^{D_C}
	p(c |\mathbf{x}^T_i;\mathbf{\theta}_T) \log p(c
	|\mathbf{x}^S_i;\mathbf{\theta}_S).
	\label{eqn:loss_ts}
\end{align}
\vspace{-15pt}

\subsection{T/S Learning with Interpolated Labels}
However, in T/S learning, the knowledge from the teacher is not accurate when the teacher's classification decision is incorrect. To deal with this, Hinton et. al. \cite{hinton2015distilling} later suggested an interpolated T/S method which uses a weighted sum of the soft posteriors and the one-hot hard label to train the student model. Assuming that the sequence of one-hot ground truth class labels that both $X^T$ and $X^S$ are aligned with is
$\mathbf{C} = \{c_1, \ldots, c_N\}$,
The interpolated T/S learning aims to minimizing the loss function below:
\begin{align}
	&\mathcal{L}_{ITS}(\mathbf{\theta}_S) = 
	-\frac{1}{N}\sum_{i=1}^N\sum_{c = 1}^{D_C}
	\left[(1-\lambda)\mathbbm{1}[c = c_i] + \lambda p(c |\mathbf{x}^T_i;\mathbf{\theta}_T)\right] \nonumber \\
	& \qquad\qquad\qquad\qquad\qquad\quad \log p(c |\mathbf{x}^S_i;\mathbf{\theta}_S),
	\label{eqn:loss_hts}
	\vspace{-10pt}
\end{align}
where $0 \le \lambda \le 1$ is the weight for the class posteriors and $\mathbbm{1}[\cdot]$ is the indicator function which equals to 1 if the condition in the squared bracket is satisfied and 0 otherwise. Note that the interpolated T/S learning becomes soft T/S when $\lambda = 1.0$ and becomes standard cross-entropy training with hard labels when $\lambda = 0.0$. Although interpolated T/S compensates for the imperfection in knowledge transfer, the linear combination of soft and hard labels destroys the correct relationships among different
classes embedded naturally in the soft class posteriors and deviates the student model parameters from the optimal direction. Moreover, the search for the best student model is subject to the heuristic tuning of  $\lambda$ between $0$ and $1$.

\section{Conditional Teacher-Student Learning}
\label{sec:smart_ts}
\begin{figure}[htpb!]
	\centering
	\includegraphics[width=0.9\columnwidth]{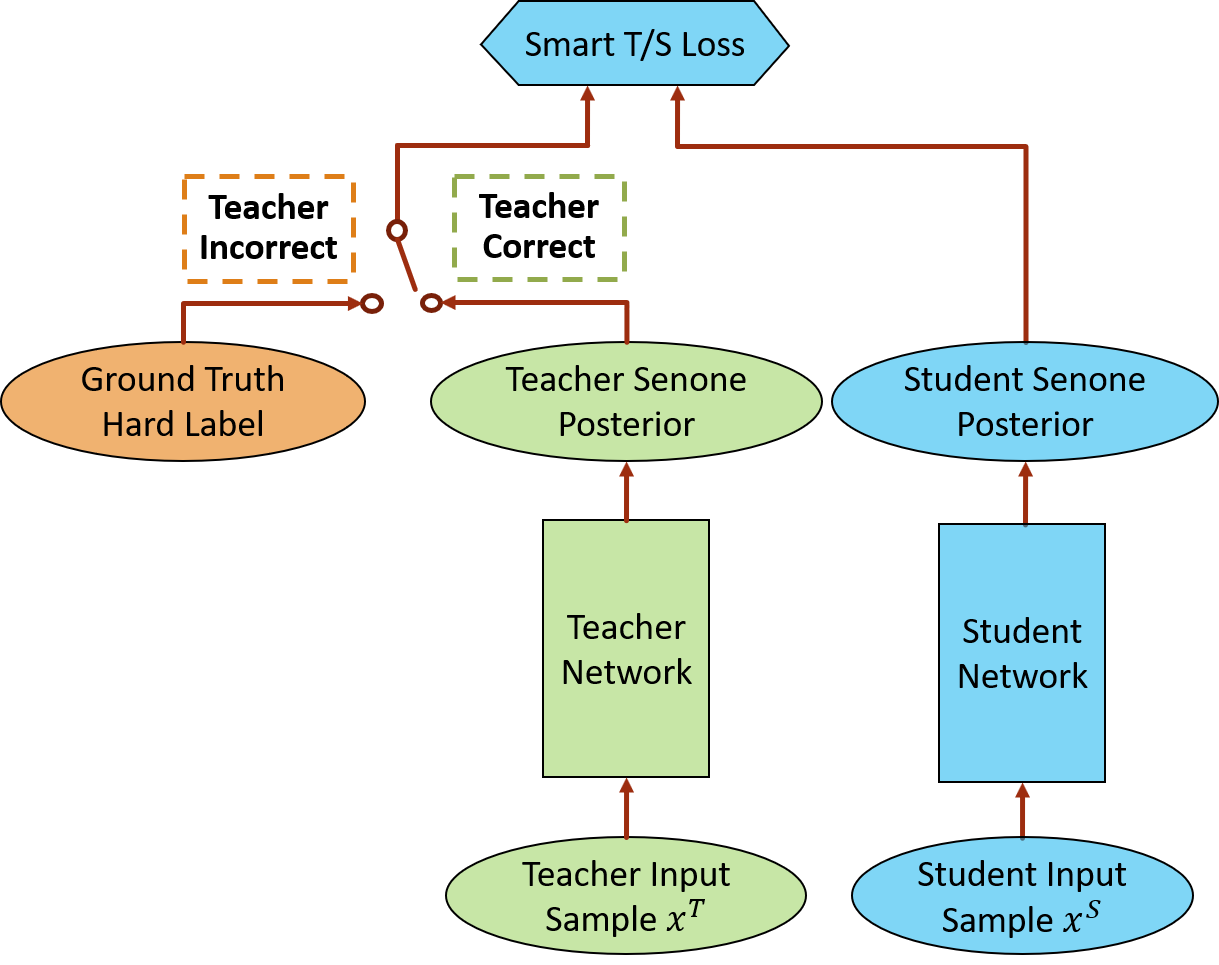}
    \vspace{-2pt}
	\caption{The framework of conditional T/S learning learning.
	}
	\label{fig:cts}
\end{figure}

Instead of blindly combining the soft and hard labels, the student network needs to be critical about the knowledge infused by the teacher network, i.e., to judge whether
the class posteriors are accurate or not before learning from them. One natural 
judgment is that the teacher's knowledge is deemed accurate when it correctly predicts the ground truth given the input samples, and deemed inaccurate otherwise. Therefore, the training target for the student model should be \emph{conditioned on} the correctness of the teacher's prediction, i.e., the student network exclusively uses the soft posteriors from the teacher as the training target when the teacher is correct and uses the hard label instead when the teacher is wrong as shown in Fig. \ref{fig:cts}. 
\vspace{-1pt}

In other words, assuming $\mathbf{Y} = \{\mathbf{y}_1, \ldots, \mathbf{y}_N\}, \mathbf{y}_i
\in \mathbbm{R}^{D_C}$ to be the sequence of \emph{conditional} class label vectors used as
the target to train the student network, the $c^{\text{th}}$ element of 
$\mathbf{y}_i$ becomes
\begin{align} \setlength\arraycolsep{-2pt}
\hspace{-5pt} y_{i,c} = \begin{cases}
	p(c | \mathbf{x}^T_i; \mathbf{\theta}_T), 
	\; \argmax_{k\in \{1, \ldots, D_C \}} p(k|
	\mathbf{x}^T_i; \mathbf{\theta}_T) = c_i \hspace{-10pt} \\
	\mathbbm{1}[c = c_i], \quad \; \text{otherwise},
\end{cases}
\label{eqn:conditional_label}
\vspace{-10pt}
\end{align}
under conditional T/S learning. That is to say, the conditional class label
$\mathbf{y}_i$ is a soft vector of class posteriors if the teacher is
correct and a hard one-hot vector if the teacher is wrong.  The loss
function to be minimized is formulated as the cross-entropy between the
conditional class labels and the class posteriors generated by the student
network as follows:
\begin{align}
	& \mathcal{L}_{CTS}(\mathbf{\theta}_S) = -\frac{1}{N} \sum_{i=1}^N \sum_{c = 1}^{D_C}
        y_{i,c}\log p(c|\mathbf{x}^S_i;\mathbf{\theta}_S) \nonumber \\
	&=-\frac{1}{N}\sum_{i=1}^N \left\{ \left[ \sum_{c = 1}^{D_C}
	p(c |\mathbf{x}^T_i;\mathbf{\theta}_T) \log p(c |\mathbf{x}^S_i;\mathbf{\theta}_S) \right] \right.
	\nonumber \\
	& \qquad\qquad\qquad\qquad\quad\mathbbm{1}[\argmax_{k\in \{1, \ldots, D_C \}} p(
	k|\mathbf{x}^T_i;\mathbf{\theta}_T) = c_i] \nonumber \\ 
	& \quad\; \left. + \log p(
				c_i| \mathbf{x}^S_i;
	\mathbf{\theta}_S) \vphantom{\sum_{c=1}^{D_C}} \mathbbm{1}[\argmax_{k\in \{1, \ldots, D_C \}} p(
k|\mathbf{x}^T_i;\mathbf{\theta}_T) \ne c_i] \right\}.
	\label{eqn:loss_cts}
\end{align}
The student network parameters are optimized through standard back
propagation with stochastic gradient decent.
With conditional T/S learning, the student can learn from only the selected \emph{accurate} knowledge generated by the teacher while simultaneously take advantage of the well-preserved probabilistic relationships among different classes and is thus expected to achieve improved performance in classification tasks.

\section{Conditional T/S Learning for Acoustic Model Adaptation}

With the advent of deep acoustic models, the performance of ASR has been greatly improved \cite{
jaitly2012application, DNN4ASR-hinton2012, deng2013recent}.
 A deep acoustic
model takes the speech frames as the input and predicts the
corresponding senone posteriors at the output layer. 
To achieve robust ASR over different domains and speakers, we apply conditional T/S learning to the domain and speaker adaptation of deep acoustic models. In these tasks, both teacher and student networks represent deep  acoustic models,  $\mathbf{X}^T$ and
$\mathbf{X}^S$ are sequences of input speech frames, and $c$ denotes one senone in the set of all possible senones $\{1, \ldots, D_C\}$ predicted by the teacher and student acoustic models. 


\subsection{Conditional T/S Learning for Domain Adaptation}
\label{sec:domain_adapt}
ASR suffers from performance degradation when a
well-trained acoustic model is applied in a new domain \cite{Li14overview}. 
T/S learning \cite{li2017large, li2018developing, watanabe2017student} and adversarial learning \cite{grl_sun, dsn_meng, meng2019aadit, meng2018cycle, meng2018afm} are two effective approaches that can suppress this domain mismatch by adapting a source-domain acoustic model to target-domain speech. T/S learning is more suited for the situation where unlabeled parallel data is available for adaptation,\footnote{The parallel data can be either recorded or simulated as in \cite{li2017large}.}
in which a sequence of source-domain speech features is fed as the input to a source-domain teacher model and a parallel sequence of target-domain features is at the input to the target-domain student  model to optimize the student model parameters by minimizing the T/S loss in Eq. \eqref{eqn:loss_ts}.   


To further improve T/S learning, we introduce the conditional T/S learning by using the ground truth hard labels $\mathbf{C}$ of the adaptation data and propose the following steps for domain adaptation.
\begin{enumerate}
    \item \label{item:da_init} Use a well-trained source-domain acoustic model as the
teacher network and initialize the student network with the parameters
of the teacher.
    \item \label{item:da_data} Use \emph{paralleled} source and target domain adaptation data as $\mathbf{X}^T$ and $\mathbf{X}^S$, respectively. All
pairs of $\mathbf{x}^T_i$ and $\mathbf{x}^S_i, \forall i \in \{1, \ldots,
N\}$ are frame-by-frame synchronized.
    \item \label{item:da_ts} Perform T/S learning \cite{li2014learning} to train the student network by minimizing $\mathcal{L}_{TS}(\mathbf{\theta}_S)$ in Eq. \eqref{eqn:loss_ts}.
    \item \label{item:da_sts} After Step \ref{item:da_ts}, the student network has performed reasonably well on target-domain data. Conduct \emph{conditional T/S} learning with conditional senone labels $\mathbf{Y}$ defined in Eq. \eqref{eqn:conditional_label} to train the student network by minimizing $\mathcal{L}_{CTS}(\mathbf{\theta}_S)$ in Eq. \eqref{eqn:loss_cts}.
    \item \label{item:decode} Use the optimized student network as the adapted acoustic model
for decoding test utterances in the target domain.
\end{enumerate}

\subsection{Conditional T/S Learning for Speaker Adaptation}
\label{sec:speaker_adapt}

Speaker adaptation aims at learning a set of speaker-dependent (SD) acoustic models by adapting an speaker-independent (SI) acoustic model to the speech of target speakers.
Different from domain adaptation, speaker adaptation has only access to 
limited adaptation data from target speakers and has no access to the source-domain data.

Many techniques have been proposed for speaker adaptation of deep acoustic models, such as
regularization-based \cite{kld_yu, l2_liao, multi_huang},
transformation-based \cite{feature_seide, lhuc_pawel_1},
singular value decomposition-based \cite{svd_xue_1,svd_zhao}, 
subspace-based \cite{sc_xue, fhl} and adversarial learning-based \cite{meng2019asa, meng2018speaker} approaches. Among these, KL divergence (KLD) regularization \cite{kld_yu} is one of the most popular methods to prevent the adapted model from overfitting the limited speaker data. 
This regularization is  realized  by augmenting the training criterion with the KLD between the output distributions of the SD and SI models.

Apparently, the KLD adaptation is a special case of the interpolated T/S learning \cite{asami_2017}, in which the SI model acts as a teacher, the SD model acts as a student, and both take the adaptation data as input. 
The teacher network is more like a regularizer that constrains the student network from straying too far away from the teacher network. However, the linear combination between soft posteriors and hard labels does not make full use of two knowledge sources, and the best regularization weight $\lambda$ is subject to heuristic tuning. 
We apply the conditional T/S learning to further improve the KLD adaptation. That is, when the SI model makes the right predictions, the SD model exclusively learns from the SI model; when the SI model is wrong, the adaptation target backs off to the hard labels. 

Note that since the SD model grows from the SI model, the adaptation can be interpreted as a self-taught learning process. In the step of learning from the SI model, the SD model basically reviews what it has already known once again, which sounds not quite informative. However, if we remove this step, i.e., adapt the SD model only when the SI model makes a mistake, the performance degrades. This is because using partial training set leads to catastrophic forgetting and skews the estimation of the senone distributions for the target speaker towards those samples the teacher model makes mistakes on and there is no guarantee that the student model can work well on those samples the teacher model is good at.

The conditional T/S learning for  speaker adaptation consists of the following steps.
\begin{enumerate}
    \item \label{item:sa_init} Use a well-trained SI acoustic model as the
teacher network and initialize the student network with the parameters
of the teacher.
    \item Use adaptation data from a target speaker as both $\mathbf{X}^T$ and $\mathbf{X}^S$. 
    \item \label{item:sa_sts} Perform \emph{conditional T/S} learning with conditional senone labels $\mathbf{Y}$ defined in Eq. \eqref{eqn:conditional_label} to train the student network by minimizing $\mathcal{L}_{CTS}(\mathbf{\theta}_S)$ in Eq. \eqref{eqn:loss_cts}.
    \item \label{item:sa_tgt} Use the optimized student network as the SD acoustic model for this target speaker.
\end{enumerate}

For unsupervised speaker adaptation, we use the SI model to generate the hard labels $\mathbf{C}$ to judge the SI model itself. 
Since the recognition hypotheses are generated through the cooperation of the SI acoustic model along with the language model, the derived hard labels are expected to be more accurate than the senone classification decisions generated by only the SI model at the frame level.

\section{Experiments}
\subsection{Domain Adaptation}
\label{sec:exp_domain_adapt}
As a major category of domain adaptation, we first verify conditional T/S
learning with environment adaptation experiments.
Specifically, we adapt a well-trained clean acoustic model to the noisy
training data of CHiME-3 \cite{chime3_barker} using different methods. The
CHiME-3 dataset incorporates Wall Street Journal (WSJ) corpus sentences
spoken in challenging noisy environments, recorded using a 6-channel
tablet.  The real far-field noisy speech from the 5th microphone channel in
CHiME-3 development data set is used for testing. A standard WSJ 5K word
3-gram language model (LM) is used for decoding.

As a source-domain acoustic model, a clean long short-term memory (LSTM)-
recurrent neural networks (RNN) \cite{sak2014long, meng2017deep, erdogan2016multi} is trained with 9137
clean training utterances of CHiME-3 dataset by using cross-entropy
criterion.  The 29-dimensional log Mel filterbank features together with
1st and 2nd order delta features (totally 87-dimensional) for both the
clean and noisy utterances are extracted by following the process in
\cite{li2012improving}. The features are fed as the input of the LSTM after
global mean and variance normalization. The LSTM has 4 hidden layers with
1024 hidden units for each layer. A 512-dimensional projection layer is
inserted on top each hidden layer to reduce the number of parameters.  The
output layer of the LSTM has 3012 output units corresponding to 3012 senone
labels. There is no frame stacking, and the output HMM senone label is
delayed by 5 frames.  Senone-level forced alignment of the clean data is
generated using a Gaussian mixture model-HMM system.
The clean CHiME-3 LSTM acoustic model
achieves 7.43\% and 38.96\% WERs on clean and real noisy test data of
CHiME-3, respectively. The clean LSTM acoustic model serves as the teacher
network in the subsequent T/S learning methods. Trained with noisy and clean
data using their one-hot hard labels, the multi-style LSTM acoustic model
achieves 19.84\% WER on the noisy test data.

\begin{table}
\centering
\begin{tabular}[c]{c|c|c|c|c|c}
	\hline
	\hline
	System & BUS & CAF & PED & STR & Avg. \\
	\hline
	Unadapted & 43.47 & 45.93 & 30.43 & 36.13 & 38.96 \\
 	\hline
 	Hard Label & 24.92 & 20.63 & 15.96 & 18.01 & 19.84 \\ 
	\hline
	Soft T/S &  22.46 & 19.10 & 14.88 & 16.47 & 18.20  \\
	\hline
	IT/S ($\lambda=0.2$) & 24.84 & 19.79 & 15.55 & 18.36 & 19.60 \\
	\hline
	IT/S ($\lambda=0.5$) & 22.61 & 18.94 & 14.52 & 18.43 & 18.59 \\
	\hline
	IT/S ($\lambda=0.8$) & 23.51 & 19.10 & 14.49 & 16.56 & 18.37 \\
	\hline
	Conditional T/S & 20.72 & 17.46 & 12.52 & 15.09 & \textbf{16.42} \\
	\hline
	\hline
	\end{tabular}
	\caption{The WER (\%) performance of environment adaptation
		using one-hot hard label, T/S,
	interpolated T/S (IT/S)  and conditional T/S learning on the real noisy test set of CHiME-3. }
\label{table:domain_chime3_wer}
\vspace{-20 pt}
\end{table}
For domain adaptation in \cite{li2017large}, parallel data consisting of 9137
pairs of clean and noisy utterances in the CHiME-3 training set are used as
the adaptation data for T/S learning. In order to make the student model
invariant to environments, the training data for student model should
include both clean and noisy data. Therefore, we extend the original T/S
learning work by also including 9137 pairs of the clean and clean
utterances in CHiME-3 for adaptation as in \cite{meng2018adversarial}.  
As shown in Table \ref{table:domain_chime3_wer}, soft T/S learning achieves
18.20\% average WERs after environment adaptation, which is 51.3\%
relative improvement over the clean model.  
To further improve the student model, we perform conditional T/S
learning with the help of hard labels as described in Section \ref{sec:domain_adapt}.  As a comparison, we
conduct interpolated T/S learning \cite{hinton2015distilling} with different weights 
for soft labels. The conditional T/S learning achieves 16.42\% average WERs
with 9.8\% and 11.7\% relative improvements over soft T/S learning and the best performed interpolated 
T/S $(\lambda = 0.5)$, respectively.

Note that we can get a better student model if we have a better teacher model. Then, we did a quick experiment by using a 375 hour-trained Cortana model which was used in \cite{li2017large} as the teacher model to learn the student model with the same CHiME-3 parallel data. The soft T/S model gets 13.56\% WER which is significantly better the one in Table 1, and the conditional T/S can reach 11.13\% WER, which stands for 17.9\% relative improvement over soft T/S.

\subsection{Speaker Adaptation}
\label{sec:exp_ts}

We further perform speaker adaptation on a Microsoft internal Windows Phone short message dictation (SMD) task. 
The test set consists of 7 speakers with a total number of 20,203 words. 
A separate adaptation set of 200 sentences per speaker is used for  model adaptation.
We train an SI LSTM acoustic model with 2600 hours of Microsoft internal live US English data.
This SI model has 4 hidden  LSTM layers with 1024 units in each layer 
and the output size of each LSTM layer is reduced to 512 by linear projection. 
The acoustic feature is  80-dimensional log Mel filterbank. The output layer has a
dimension of 5980. The LSTM-RNN is trained to minimize the frame-level
cross-entropy criterion. There is no frame stacking, and the output HMM
state label is delayed by 5 frames. A trigram LM is used for decoding with
around 8M n-grams. This SI LSTM acoustic model achieves 13.95\% WER on the
SMD test set. 

We perform conditional T/S learning as in Section
\ref{sec:speaker_adapt} to adapt the SI LSTM with 200 utterances in the 
adaptation set for each test speaker. 
For supervised adaptation, the hard labels come from the human transcription though forced alignment. For unsupervised adaptation, we use the SI model to generate the hypothesis. 
As a comparison, the standard adaptation with hard labels and KLD
adaptation \cite{kld_yu} with regularization weights $\lambda$ of $0.2, 0.5$ and $0.8$
are also conducted to adapt the SI LSTM. Note that the adaptation with hard labels is equivalent to KLD adaptation with $\lambda=0$.
As in Table \ref{table:speaker_adapt_wer}, the KLD adaptation produces its best WERs of 12.54\% and 13.55\% for supervised and unsupervised adaptation at $\lambda = 0.5$, respectively. 
The conditional T/S learning outperforms the KLD adaptation. It achieves 12.17\% WER for supervised adaptation, which is 12.8\%
and 3.0\% relative gain over the SI model and the best performed KLD adaptation 
$(\lambda = 0.5)$. For unsupervised adaptation, the conditional T/S learning achieves
13.21\% WER, which is 5.3\% and 2.5\% relative gain over the SI acoustic
model and KLD adaptation.

\begin{table}
\centering
\begin{tabular}[c]{c|c|c}
	\hline
	\hline
	System & Supervised & Unsupervised \\
	\hline
	SI & \multicolumn{2}{c}{13.95} \\
	\hline
	Hard Label & 13.20 & 13.77 \\
	\hline
	KLD ($\lambda = 0.2)$ & 12.61 & 13.65 \\ 
	\hline
	KLD $(\lambda = 0.5)$ & 12.54 & 13.55 \\ 
	\hline
	KLD $(\lambda = 0.8)$ & 13.17 & 13.72 \\ 
	\hline
	Conditional T/S & \textbf{12.17} & \textbf{13.21} \\
	\hline
	\hline
	\end{tabular}
	\caption{The WER (\%) performance of speaker adaptation
		using one-hot hard label, KLD 
	and conditional T/S learning on Microsoft SMD task. The SI LSTM model is trained with
2600 hours Microsoft live US English data.}
\label{table:speaker_adapt_wer}
\vspace{-12pt}
\end{table}





\section{Conclusion}
We proposed a conditional T/S learning method, in which the student network
selectively learns from either the soft posteriors generated by the
teacher network or the one-hot hard label \emph{conditioned on} whether the teacher
makes correct decisions or not. 
Instead of blindly following whatever knowledge the teacher infuses as in the conventional T/S
learning, the conditional T/S learning pursues the most trustworthy knowledge throughout the training, eliminating the burden tuning interpolation weights.
We applied conditional T/S learning to domain adaptation and obtained 9.8\%
relative WER improvement over a strong T/S learning baseline on the CHiME-3 dataset. 
For speaker adaptation, the conditional T/S learning outperformed the KLD adaptation, which is equivalent to the interpolated T/S learning. It achieved 12.8\% and 5.3\% relative WER gains for supervised and unsupervised adaptations, respectively, over a well-trained SI LSTM model. 

\vfill\pagebreak

\bibliographystyle{IEEEbib}
\bibliography{refs}

\begin{thebibliography}{10}

\bibitem{li2014learning}
J.~Li, R.~Zhao, J.-T. Huang, and Y.~Gong,
\newblock ``Learning small-size {DNN} with output-distribution-based
  criteria.,''
\newblock in {\em Proc. INTERSPEECH}, 2014, pp. 1910--1914.

\bibitem{hinton2015distilling}
G.~E. Hinton, O.~Vinyals, and J.~Dean,
\newblock ``Distilling the knowledge in a neural network,''
\newblock {\em CoRR}, vol. abs/1503.02531, 2015.

\bibitem{li2017large}
J.~Li, M.~L Seltzer, X.~Wang, et~al.,
\newblock ``Large-scale domain adaptation via teacher-student learning,''
\newblock in {\em INTERSPEECH}, 2017.

\bibitem{meng2018adversarial}
Zhong Meng, Jinyu Li, Yifan Gong, and Biing-Hwang~(Fred) Juang,
\newblock ``Adversarial teacher-student learning for unsupervised domain
  adaptation,''
\newblock in {\em Proc. ICASSP}, 2018.

\bibitem{movsner2019improving}
L.~Mo{\v{s}}ner, M.~Wu, A.~Raju, et~al.,
\newblock ``Improving noise robustness of automatic speech recognition via
  parallel data and teacher-student learning,''
\newblock {\em arXiv preprint arXiv:1901.02348}, 2019.

\bibitem{kim2016sequence}
Yoon Kim and Alexander~M. Rush,
\newblock ``Sequence-level knowledge distillation,''
\newblock in {\em EMNLP}, 2016, pp. 1317--1327.

\bibitem{chen2017teacher}
Yun Chen, Yang Liu, Yong Cheng, and Victor~O.K. Li,
\newblock ``A teacher-student framework for zero-resource neural machine
  translation,''
\newblock in {\em Proc. ACL}, 2017, pp. 1925--1935.

\bibitem{li2018developing}
J.~Li, R.~Zhao, Z.~Chen, et~al.,
\newblock ``Developing far-field speaker system via teacher-student learning,''
\newblock {\em arXiv preprint arXiv:1804.05166}, 2018.

\bibitem{watanabe2017student}
S.~Watanabe, T.~Hori, J.~Le~Roux, et~al.,
\newblock ``Student-teacher network learning with enhanced features,''
\newblock in {\em Proc. ICASSP}, 2017.

\bibitem{cui2017knowledge}
J.~Cui, B.~Kingsbury, B.~Ramabhadran, et~al.,
\newblock ``Knowledge distillation across ensembles of multilingual models for
  low-resource languages,''
\newblock in {\em Proc. ICASSP}. IEEE, 2017.

\bibitem{tang2016recurrent}
Z.~Tang, D.~Wang, and Z.~Zhang,
\newblock ``Recurrent neural network training with dark knowledge transfer,''
\newblock in {\em Proc. ICASSP}, 2016.

\bibitem{asami_2017}
T.~Asami, R.~Masumura, Y.~Yamaguchi, et~al.,
\newblock ``Domain adaptation of dnn acoustic models using knowledge
  distillation,''
\newblock in {\em Proc. ICASSP}, 2017, pp. 5185--5189.

\bibitem{chan2015transferring}
W.~Chan, N.~R. Ke, and I.~Lane,
\newblock ``Transferring knowledge from a rnn to a dnn,''
\newblock in {\em INTERSPEECH}, 2015.

\bibitem{tan2018knowledge}
T.~Tan, Yanmin Q., and D.~Yu,
\newblock ``Knowledge transfer in permutation invariant training for
  single-channel multi-talker speech recognition,''
\newblock in {\em ICASSP}, 2018.

\bibitem{lu2017knowledge}
L.~Lu, M.~Guo, and S.~Renals,
\newblock ``Knowledge distillation for small-footprint highway networks,''
\newblock in {\em ICASSP}, 2017.

\bibitem{jaitly2012application}
N.~Jaitly, P.~Nguyen, A.~Senior, and V.~Vanhoucke,
\newblock ``Application of pretrained deep neural networks to large vocabulary
  speech recognition,''
\newblock in {\em Proc. INTERSPEECH}, 2012.

\bibitem{DNN4ASR-hinton2012}
G.~Hinton, L.~Deng, D.~Yu, et~al.,
\newblock ``Deep neural networks for acoustic modeling in speech recognition:
  The shared views of four research groups,''
\newblock {\em IEEE Signal Processing Magazine}, vol. 29, no. 6, pp. 82--97,
  2012.

\bibitem{deng2013recent}
Li~Deng, Jinyu Li, Jui-Ting Huang, et~al.,
\newblock ``Recent advances in deep learning for speech research at
  {Microsoft},''
\newblock in {\em ICASSP}, 2013.

\bibitem{Li14overview}
J.~Li, L.~Deng, Y.~Gong, et~al.,
\newblock ``An overview of noise-robust automatic speech recognition,''
\newblock {\em IEEE/ACM Transactions on Audio, Speech and Language Processing},
  2014.

\bibitem{grl_sun}
S.~Sun, B.~Zhang, L.~Xie, and Y.~Zhang,
\newblock ``An unsupervised deep domain adaptation approach for robust speech
  recognition,''
\newblock {\em Neurocomputing}, 2017.

\bibitem{dsn_meng}
Z.~Meng, Z.~Chen, V.~Mazalov, J.~Li, and Y.~Gong,
\newblock ``Unsupervised adaptation with domain separation networks for robust
  speech recognition,''
\newblock in {\em Proceeding of ASRU}, 2017.

\bibitem{meng2019aadit}
Z.~Meng, J.~Li, and Y.~Gong,
\newblock ``Attentive adversarial learning for domain-invariant training,''
\newblock in {\em Proc. ICASSP}, 2019.

\bibitem{meng2018cycle}
Z.~Meng, J.~Li, and Y.~Gong,
\newblock ``Cycle-consistent speech enhancement,''
\newblock {\em Interspeech}, 2018.

\bibitem{meng2018afm}
Z.~Meng, J.~Li, and Y.~Gong,
\newblock ``Adversarial feature-mapping for speech enhancement,''
\newblock {\em Interspeech}, 2018.

\bibitem{kld_yu}
D.~Yu, K.~Yao, H.~Su, et~al.,
\newblock ``Kl-divergence regularized deep neural network adaptation for
  improved large vocabulary speech recognition,''
\newblock in {\em Proc. ICASSP}, 2013.

\bibitem{l2_liao}
H.~Liao,
\newblock ``Speaker adaptation of context dependent deep neural networks,''
\newblock in {\em Proc. ICASSP}, May 2013, pp. 7947--7951.

\bibitem{multi_huang}
Z.~Huang, J.~Li, S.~Siniscalchi, et~al.,
\newblock ``Rapid adaptation for deep neural networks through multi-task
  learning,''
\newblock in {\em Interspeech}, 2015.

\bibitem{feature_seide}
F.~Seide, G.~Li, X.~Chen, and D.~Yu,
\newblock ``Feature engineering in context-dependent deep neural networks for
  conversational speech transcription,''
\newblock in {\em Proc. ASRU}, Dec 2011, pp. 24--29.

\bibitem{lhuc_pawel_1}
P.~Swietojanski, J.~Li, and S.~Renals,
\newblock ``Learning hidden unit contributions for unsupervised acoustic model
  adaptation,''
\newblock {\em IEEE/ACM Transactions on Audio, Speech, and Language
  Processing}, vol. 24, no. 8, pp. 1450--1463, Aug 2016.

\bibitem{svd_xue_1}
J.~Xue, J.~Li, and Y.~Gong,
\newblock ``Restructuring of deep neural network acoustic models with singular
  value decomposition.,''
\newblock in {\em Interspeech}, 2013, pp. 2365--2369.

\bibitem{svd_zhao}
Y.~Zhao, J.~Li, and Y.~Gong,
\newblock ``Low-rank plus diagonal adaptation for deep neural networks,''
\newblock in {\em Proc. ICASSP}, 2016.

\bibitem{sc_xue}
S.~Xue, O.~Abdel-Hamid, H.~Jiang, et~al.,
\newblock ``Fast adaptation of deep neural network based on discriminant codes
  for speech recognition,''
\newblock {\em IEEE/ACM Transactions on Audio, Speech, and Language
  Processing}, Dec 2014.

\bibitem{fhl}
L.~Samarakoon and K.~C. Sim,
\newblock ``Factorized hidden layer adaptation for deep neural network based
  acoustic modeling,''
\newblock {\em IEEE/ACM Transactions on Audio, Speech, and Language
  Processing}, vol. 24, no. 12, pp. 2241--2250, Dec 2016.

\bibitem{meng2019asa}
Zhong Meng, Jinyu Li, and Yifan Gong,
\newblock ``Adversarial speaker adaptation,''
\newblock in {\em Proc. ICASSP}, 2019.

\bibitem{meng2018speaker}
Z.~Meng, J.~Li, Z.~Chen, et~al.,
\newblock ``Speaker-invariant training via adversarial learning,''
\newblock in {\em Proc. ICASSP}, 2018.

\bibitem{chime3_barker}
J.~Barker, R.~Marxer, E.~Vincent, and S.~Watanabe,
\newblock ``The third {CHiME} speech separation and recognition challenge:
  Dataset, task and baselines,''
\newblock in {\em Proc. ASRU}, 2015, pp. 504--511.

\bibitem{sak2014long}
F.~Beaufays H.~Sak, A.~Senior,
\newblock ``Long short-term memory recurrent neural network architectures for
  large scale acoustic modeling,''
\newblock in {\em Interspeech}, 2014.

\bibitem{meng2017deep}
Z.~Meng, S.~Watanabe, J.~R. Hershey, and H.~Erdogan,
\newblock ``Deep long short-term memory adaptive beamforming networks for
  multichannel robust speech recognition,''
\newblock in {\em ICASSP}, 2017.

\bibitem{erdogan2016multi}
H.~Erdogan, T.~Hayashi, J.~R. Hershey, et~al.,
\newblock ``Multi-channel speech recognition: Lstms all the way through,''
\newblock in {\em CHiME-4 workshop}, 2016, pp. 1--4.

\bibitem{li2012improving}
J.~Li, D.~Yu, J.-T. Huang, and Y.~Gong,
\newblock ``Improving wideband speech recognition using mixed-bandwidth
  training data in {CD-DNN-HMM},''
\newblock in {\em Proc. SLT}. IEEE, 2012, pp. 131--136.

\end{thebibliography}

\end{document}